\documentclass[twoside]{article}

\usepackage[accepted]{aistats2025}

\usepackage{graphicx}

\usepackage[round]{natbib}

\usepackage{url}
\usepackage{booktabs}
\usepackage{multirow}
\usepackage{tikz}
\usepackage{wrapfig}
\usetikzlibrary{shapes.geometric} 
\usetikzlibrary{shapes.geometric, shadows.blur, positioning}
\usepackage{amsmath}

\definecolor{camblue}{RGB}{0, 115, 207}
\definecolor{camred}{RGB}{214, 8, 59}
\definecolor{camyellow}{RGB}{227, 114, 34}
\definecolor{camgreen}{RGB}{88, 166, 24}
\definecolor{campurple}{RGB}{142, 37, 141}
\definecolor{camindigo}{RGB}{0, 179, 190}
\definecolor{camdeepblue}{RGB}{0, 62, 114}
\definecolor{camdeepred}{RGB}{114, 28, 59}
\definecolor{camdeeppurple}{RGB}{65, 45, 93}

\newcommand{\poutp}{p_{\text{out}}^{\text{proxy}}}

\begin{document}

%

%
\runningauthor{Andi Zhang, Tim Z. Xiao, Weiyang Liu, Robert Bamler, Damon Wischik }

\twocolumn[

\aistatstitle{Your Finetuned Large Language Model is Already a Powerful Out-of-distribution Detector}

\aistatsauthor{ Andi Zhang$^{*,1,3}$~~~~Tim Z. Xiao$^{2,4,5}$~~~~Weiyang Liu$^{3,5}$~~~~Robert Bamler$^{2}$~~~~Damon Wischik$^{3}$ }

\vspace{0.5em}

\aistatsaddress{ $^1$University of Manchester ~~~~ $^2$University of T\"ubingen ~~~~ $^3$University of Cambridge \\ $^4$IMPRS-IS ~~~~ $^5$Max Planck Institute for Intelligent Systems, T\"ubingen\\ $^*$Correspondence to: \texttt{az381@cantab.ac.uk}} ]

\begin{abstract}
We revisit the likelihood ratio between a pretrained large language model (LLM) and its finetuned variant as a criterion for out-of-distribution (OOD) detection. The intuition behind such a criterion is that, the pretrained LLM has the prior knowledge about OOD data due to its large amount of training data, and once finetuned with the in-distribution data, the LLM has sufficient knowledge to distinguish their difference. Leveraging the power of LLMs, we show that, the likelihood ratio can serve as an effective OOD detection criterion. Moreover, we apply the proposed LLM-based likelihood ratio to detect OOD questions in question-answering (QA) systems, which can be used to improve the performance of specialized LLMs for general questions. Given that likelihood can be easily obtained by the loss functions within contemporary neural network frameworks, it is straightforward to implement this approach in practice. Since both the pretrained LLMs and its various finetuned models are widely available from online platforms such as Hugging Face, our proposed criterion can be effortlessly incorporated for OOD detection without the need for further training. We conduct comprehensive evaluation across on multiple settings, including far OOD, near OOD, spam detection, and QA scenarios, to demonstrate the effectiveness of the method. Code can be found at \url{https://github.com/andiac/LLMOODratio}
\end{abstract}

\section{INTRODUCTION}
Detecting out-of-distribution (OOD) is crucial for the safety of artificial intelligence systems. OOD detection aims to identify inputs that substantially deviate from the training data the model was trained on, ensuring the system is alerted to these discrepancies. This capability to identify OOD and anomalous data is particularly critical in high-stakes domains such as healthcare and autonomous driving, where the stakes for accuracy and reliability are exceptionally high.

In OOD detection \citep{hendrycks2016baseline}, the term ``in-distribution'' refers specifically to the distribution of the training data. 
In natural language processing, OOD detection has been studied in \emph{small and task-specific models} for settings such as translation \citep{xiao2020wat} and question answering \citep{lyu2020you}.
With the advancement of \emph{large and general models} like large language models (LLMs), the scope of training data has significantly broadened, positioning these large models embodied with general knowledge and intelligence as ``base models''.
In the era of large foundation models~\citep{bommasani2021opportunities}, the prevalent training paradigm has shifted from  end-to-end learning towards finetuning a pretrained base model. This shift calls for a revisiting of the definition of ``in-distribution'' and its relationship to the training distribution of base models.

There is usually no prior information about the OOD data in conventional OOD detection~\citep{hendrycks2016baseline}. \citet{bishop1994novelty} introduced the idea that OOD data could be viewed as data coming from a distinct OOD distribution, suggesting the use of the likelihood ratio between this OOD distribution and the in-distribution as a detection criterion. However, given the absence of information about OOD data, \citet{zhang2022falsehoods} introduced the concept of OODProxy, a conceptual framework where the OOD distribution is represented by a proxy, incorporating what is known or assumed about the OOD data. From this perspective, it is commonly recognized that many OOD detection methods leveraging likelihood ratios are essentially applying their unique OOD proxy distributions, each reflecting different assumptions or prior knowledge concerning the nature of the OOD data~\citep{ren2019likelihood,serra2019input, schirrmeister2020understanding, zhang2021out}.

In our paper, we revisit\footnote{\citet{jin2022towards} found this method inferior to contrastive learning with smaller models. Our work shows it outperforms as model scale increases (Section~\ref{sec:relatedwork}).} that pretrained base models can function as a repository of prior knowledge for OOD data relative to in-distribution data, effectively acting as OOD proxy distributions. Guided by this insight, we discover that the likelihood ratio between the base model and its finetuned counterpart serves as an effective criterion for detecting OOD data. Moreover, for LLM-based question-answering (QA) systems, the same likelihood ratio excels in detecting OOD questions. By identifying and rejecting these OOD questions, we can greatly enhance the robustness of current QA systems.

The convenience of obtaining likelihood from loss functions in current neural networks enables a simple and straightforward implementation of this method in practice. Moreover, it is worth mentioning that numerous practitioners likely already have both a pretrained and a finetuned LLM at their disposal, with both types of models widely accessible from platforms like Hugging Face. This setup inherently equips them with the capacity for OOD detection without necessitating additional training efforts.

\section{BACKGROUND AND PRELIMINARIES}

\paragraph{OOD Detection} We start with an in-distribution dataset, denoted as $\mathcal{D}_{\text{in}}$, assuming that the data within $\mathcal{D}_{\text{in}}$ is sampled from an in-distribution probability distribution $p_{\text{in}}$. The objective of OOD detection is to determine whether a given input data $x$ originates from $p_{\text{in}}$. \citet{hendrycks2016baseline} were the first to highlight the significance of this problem in deep learning era and introduced a practical benchmark for evaluation. This involves training a model on $\mathcal{D}_\text{in}^\text{train}$, the training subset of the in-distribution dataset, with the model providing a detection criterion $S$ for identifying OOD data. We then gather a dataset from domains different from $\mathcal{D}_{\text{in}}$, labeled as $\mathcal{D}_{\text{out}}$. The effectiveness of the criterion $S$ is assessed by applying it to data from $\mathcal{D}_{\text{in}}^\text{test} \cup \mathcal{D}_{\text{out}}^\text{test}$ and evaluating the performance using metrics such as AUROC, AUPR, and FPR95 \citep{yang2022openood}. These metrics help determine how well $S$ can differentiate between the in-distribution dataset and the OOD dataset. A high performance across these metrics signifies a robust OOD detection capability.

\paragraph{``Supervised'' and ``Unsupervised'' OOD Detection}
In \citet{hendrycks2016baseline}'s foundational study, it is assumed that for the in-distribution dataset $\mathcal{D}_\text{in}$, each data point is accompanied by a classification label. This is why this problem setting is referred to as supervised OOD detection\footnote{A possible ambiguity here is that some might think having OOD data is what makes it supervised OOD detection; here we emphasize that ``supervised'' and ``unsupervised'' refer to the labels of in-distribution data.}. Following Hendrycks' work, numerous supervised OOD detection methods based on classifiers have been proposed \citep{lee2018simple, liu2020energy, wang2022vim, sun2022out, zhu2022boosting}. In contrast, our work focuses on a different scenario, as we operate under the assumption that no labels are available. This scenario is often referred to as ``unsupervised OOD detection'' or ``OOD detection without in-distribution labels''.

\paragraph{The Paradox in Unsupervised OOD Detection}
In the context of unsupervised OOD detection, \citet{nalisnick2018deep} revisit \citet{bishop1994novelty}'s suggestion that a probabilistic generative model $p_\theta$ could model the in-distribution $p_{\text{in}}$, proposing to use the model output $S(x) = p_\theta(x)$ as a criterion for evaluating a given input $x$. Surprisingly, Nalisnick et al. discovered that in certain scenario - such as when $\mathcal{D}_{\text{in}}$ is CIFAR10 and $\mathcal{D}_{\text{out}}$ is SVHN, or $\mathcal{D}_{\text{in}}$ is FashionMNIST and $\mathcal{D}_{\text{out}}$ is MNIST --- the OOD data received higher $p_\theta(x)$ scores than the in-distribution data. This counterintuitive finding has been labeled as a ``paradox''.

\paragraph{OOD Proxy}
\label{sec:oodproxy}
To address the paradox, the studies \citep{ren2019likelihood,serra2019input, schirrmeister2020understanding, zhang2021out, caterini2022entropic} have put forward the idea of utilizing the likelihood ratio as the criterion for identifying OOD data. \citet{zhang2022falsehoods} integrates these techniques into a comprehensive structure termed the OOD proxy framework. Within this framework, it is posited that in-distribution data can be characterized as samples from a distribution \(p_\text{in}\), whereas OOD data are samples from a distribution \(p_\text{out}\). According to the Neyman-Pearson lemma, the likelihood ratio represents the optimal criterion for OOD detection in theory, which is mathematically expressed as:
\begin{equation*}
    S(x) = \frac{p_\text{out}(x)}{p_\text{in}(x)}
\end{equation*}
where obtaining \(p_\text{out}\) is challenging. To address this, \citet{zhang2022falsehoods} introduced the concept of utilizing a proxy distribution, \(\poutp\), which incorporates human subjective understanding of the OOD distribution. For instance, \citet{ren2019likelihood} define \(\poutp\) as the distribution representing background statistics; \citet{serra2019input} consider it as the distribution of data compression; \citet{schirrmeister2020understanding} describe \(\poutp\) as a general distribution; and for \citet{zhang2021out}, \(\poutp\) corresponds to the distribution of a local autoregressive model.

In the OOD proxy framework, the use of likelihood as a detection criterion, following the approach of \citet{bishop1994novelty} and \citet{nalisnick2018deep}, essentially assumes a uniform distribution for \(\poutp\). The suboptimal performance associated with this method is not surprising; it highlights the limitations of an improper prior assumption.

\paragraph{Likelihood of Autoregressive Language Models}

Autoregressive language models \citep{brown2020language} are types of probabilistic models that compute the likelihood of a sentence \(x = x_1, \dots, x_T\), where \(T\) denotes the sentence length and \(x_t\) represents each word at position \(t\). By 
 definition of conditional probability, the likelihood \(p(x) = p(x_1, \dots, x_T)\) can be calculated as the product of conditional probabilities: \(p(x_1) p(x_2 | x_1) p(x_3 | x_1, x_2)\dots p(x_T | x_1, \dots, x_{T-1})\). Each term \(p(x_t | x_{<t})\) represents the probability of predicting the subsequent word \(x_t\) given all the previous words \(x_1, \dots, x_{t-1}\). In neural language models, these conditional probabilities are typically modeled using a softmax function over the output vocabulary.
 

When training or finetuning a language model on a dataset, the objective is to maximize the likelihood of the training data. This is equivalent to minimizing the negative log-likelihood, which is the cross-entropy loss between the predicted word probabilities and the true word labels at each position. Most modern neural network libraries provide built-in functions to compute this cross-entropy loss, making it straightforward to optimize the model parameters to minimize the negative log-likelihood and thus maximize the likelihood of the training data.

\section{PRETRAINED LLM AS A OOD PROXY}
\label{sec:llmasproxy}
Considering an autoregressive language model as a distribution \(p\), denote \(p_\theta\) as the pretrained large language model with parameters \(\theta\). Given an in-distribution dataset \(\mathcal{D}_\text{in}\), the model finetuned on \(\mathcal{D}_\text{in}\) is represented with parameters \(\theta'\). For an input \(x\), we introduce the out-of-distribution detection criterion \(S\) as follows:
\begin{equation}
\label{eq:lr}
    S(x) = \frac{p_\theta(x)}{p_{\theta'}(x)}.
\end{equation}
This criterion essentially employs the pretrained large language model as an OOD proxy introduced in Section~\ref{sec:oodproxy}. This strategy is particularly practical given the widespread availability of pretrained LLMs. Finetuning these models to adapt their distribution for specific domain contexts is a standard practice, meaning many practitioners may already possess finetuned LLMs that represent the distribution of their specific datasets. With both pretrained and finetuned models at hand, calculating the likelihood ratio becomes straightforward, eliminating the need for additional training. For example, suppose we possess a LLM that has been finetuned on legal documents. Given a new document \(x\), we can determine whether it is a legal document by utilizing the likelihood ratio \(S(x)\).

\begin{figure}[htbp]
\begin{center}
\begin{tikzpicture}[scale=0.85, every node/.style={scale=0.85}]
    Any permutation of the characters
   \node[ellipse, thick, draw=camblue, fill=camblue!5, minimum width=9cm, minimum height=4cm, label={[xshift=0cm, yshift=-0.9cm]above:Any permutation of the characters}] (A) at (0,-0.5) {};

    Human language
   \node[ellipse, thick, draw=camgreen, fill=camgreen!5, minimum width=7cm, minimum height=3cm, label={[xshift=0cm, yshift=-0.9cm]above:Human language}] (B) at (0,-0.9) {};

    Sentences in a specific domain
   \node[ellipse, thick, draw=camred, fill=camred!5, minimum width=5cm, minimum height=2cm, label={[xshift=0cm, yshift=-1.3cm]above:Sentences in a specific domain}] (C) at (0,-1.3) {};
\end{tikzpicture}
\end{center}
\caption{Relationship among sentences within a specific domain, the comprehensive set of human language, and all conceivable character permutations. \label{fig:venn}}
\end{figure}

Utilizing a pretrained LLM as an OOD proxy stems from the rationale that the OOD proxy distribution should encapsulate general characteristics of OOD data, including prior knowledge or subjective insights about the OOD data. While it's conceivable to presume OOD data emanate from a uniform distribution devoid of any prior knowledge, evidence presented by \citet{nalisnick2018deep} indicates that relying on a uniform distribution as the OOD proxy can be ineffective in practice. In the context of language models, the well-known infinite monkey theorem\footnote{The infinite monkey theorem states that a monkey randomly typing on a typewriter for an infinitely long period will almost surely produce any given text, such as the complete works of William Shakespeare.} \citep{borel1913mecanique, eddington2019nature} suggests that any text could theoretically be generated from a uniform distribution; however, this does not serve as an effective representation of any coherent language. Figure~\ref{fig:venn} depicts the relationship among sentences within a specific domain, the comprehensive set of human language, and all conceivable character permutations. In fact, coherent human language forms a minor subset of all potential character arrangements. Consequently, the assumption that OOD data ought to represent meaningful human language constitutes a strong prior.

Therefore, we advocate for the use of a pretrained LLM as a more suitable OOD proxy. Given their extensive parameters and training on vast corpora (for example, the Llama-2 model, as mentioned by \citet{touvron2023llama}, is trained on 7 trillion tokens), it is plausible to consider that an LLM encompasses the breadth of human language. Assuming the LLM estimates the distribution of all human language, it is logical to designate the pretrained LLM as the OOD proxy.

\section{LIKELIHOOD RATIO OOD DETECTION FOR QA SYSTEMS}
\label{sec:qa}
\begin{figure}[htbp]
\begin{center}
\begin{tikzpicture}[scale=1, every node/.style={scale=1}]
\node[inner sep=0pt, anchor=north] (prompt) at (0,0)
    {\includegraphics[page=1, width=0.23\textwidth]{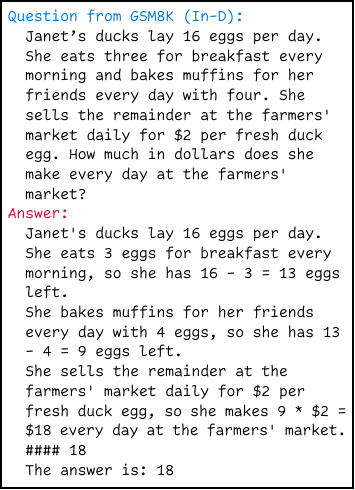}};
\node[inner sep=0pt, anchor=north] (prompt) at (4,0)
    {\includegraphics[page=2, width=0.23\textwidth]{prompt.pdf}};
\end{tikzpicture}
\end{center}
\caption{Example question-answer sets produced by MetaMath-7B. The responses to In-D questions are accurate and logical. However, for OOD questions, MetaMath-7B generates unreasonable answers, responding to a straightforward query with unnecessary mathematical calculations or producing repetitive sentences with no useful information. For the complete image, please see Appendix. \label{fig:qaexample}}
\end{figure}

In question-answering (QA) systems, identifying OOD questions is crucial for enhancing system robustness through their rejection. However, detecting OOD questions is challenging due to the often brief and uninformative nature of the questions submitted to QA systems, rendering the direct application of the likelihood ratio on the questions themselves ineffective for OOD detection. To overcome this issue, we leverage the observation that while a finetuned LLM generates pertinent answers to in-distribution questions, it tends to produce unreasonable sentences in response to OOD questions (Figure~\ref{fig:qaexample}). Therefore, we propose a novel approach: for each question, we have the finetuned LLM generate an answer, and then we apply an OOD detection criterion specifically designed for the question-answer pair.

Formally, in the context of autoregressive large language models, consider a question $q = q_1,\dots, q_{T_q}$, from which we generate an answer $a = a_1,\dots, a_{T_a}$ by sampling from the conditional distribution $p(\cdot|q)$. We define the following criterion:
\begin{align*}
    S_q(q, a) &= \frac{p_\theta(q)}{p_{\theta'}(q)},\\
    S_a(q, a) &= \frac{p_\theta(a)}{p_{\theta'}(a)},\\
    S_{q,a}(q, a) &= \frac{p_\theta(q, a)}{p_{\theta'}(q, a)},\\
    S_{a|q}(q,a) &= \frac{p_\theta(a | q)}{p_{\theta'}(a | q)} = \frac{p_\theta(q, a)p_{\theta'}(q)}{p_{\theta'}(q,a)p_\theta(q)} \\
    &= \frac{S_{q,a}(q, a)}{S_q(q, a)},
\end{align*}
where $S_q$, $S_a$, $S_{q,a}$, and $S_{a|q}$ are defined as follows: $S_q$ is the likelihood ratio for the question, $S_a$ for the answer, $S_{q,a}$ for the question-answer pair, and $S_{a|q}$ for the answer given the question. All these criterion are the ratios between the likelihoods assigned by the finetuned model to those assigned by the base model. It is difficult to determine intuitively which of these four criteria performs best. In the following experimental section, we will empirically demonstrate which criterion is more effective in practice.

\begin{table*}[t]
\scriptsize
\setlength{\tabcolsep}{12pt}
\renewcommand{\arraystretch}{1.05}
\centering
\begin{tabular}{cclccc}
\specialrule{0em}{0pt}{-10pt}
\toprule
\multicolumn{1}{c}{\bf In-D} &\multicolumn{1}{c}{\bf OOD} &\multicolumn{1}{c}{\bf Method} &\multicolumn{1}{c}{\bf AUROC $\uparrow$} &\multicolumn{1}{c}{\bf AUPR (OOD) $\uparrow$} &\multicolumn{1}{c}{\bf FPR95 $\downarrow$}\\
\midrule
\multirow{33}{*}{20NG}  & \multirow{11}{*}{SST-2}     & \citet{zhou2021contrastive} & 0.978 & 0.865 & 0.015  \\
      &            & CE \citep{hendrycks2016baseline} & 0.981 & 0.942  & 0.087 \\
      &            & TAPT \citep{gururangan2020don} & 0.981 & 0.939 & 0.088 \\
      &            & SupCon \citep{khosla2020supervised} & 0.980 & 0.943 & 0.094 \\
      &            & \citet{uppaal2023fine} & \textbf{1.000} & 0.999  & \textbf{0.000}\\
      &            & Llama-7B LH & 0.008 & 0.541 & 0.999  \\
      &            & Llama-7B LR & \textbf{1.000} & \textbf{1.000} & \textbf{0.000} \\
      &            & Mistral-7B LH  & 0.008 & 0.541 & 1.000 \\
      &            & Mistral-7B LR & 0.995 & 0.999 & 0.009 \\
      &            & Llama-13B LH & 0.009 & 0.541 & 1.000  \\
      &            & Llama-13B LR & \textbf{1.000} & \textbf{1.000} & \textbf{0.000} \\
             \cmidrule(l){2-6}
      & \multirow{11}{*}{RTE}        & \citet{zhou2021contrastive} &0.956  & 0.860 & 0.312 \\
      &            & CE \citep{hendrycks2016baseline} &0.945  & 0.902 & 0.285 \\
      &            & TAPT \citep{gururangan2020don} &0.919 &  0.869 & 0.352 \\
      &            & SupCon \citep{khosla2020supervised} &0.952 &  0.914 & 0.248 \\
      &            & \citet{uppaal2023fine} & \textbf{1.000} &  0.999 & \textbf{0.000} \\
      &            & Llama-7B LH  & 0.063 & 0.443 & 0.998 \\
      &            & Llama-7B LR & \textbf{1.000} & \textbf{1.000} & 0.001 \\
      &            & Mistral-7B LH & 0.074 & 0.446 & 0.998 \\
      &            & Mistral-7B LR & 0.997 & 0.999 & 0.006  \\
      &            & Llama-13B LH & 0.070 & 0.445 & 0.997 \\
      &            & Llama-13B LR & \textbf{1.000} & \textbf{1.000} & \textbf{0.000}  \\
             \cmidrule(l){2-6}
      & \multirow{11}{*}{IMDB}       & \citet{zhou2021contrastive} & 0.969 & 0.996 & 0.144 \\
      &            & CE \citep{hendrycks2016baseline} &0.961 & 0.995 & 0.206 \\
      &            & TAPT \citep{gururangan2020don} & 0.965 & 0.995 & 0.159 \\
      &            & SupCon \citep{khosla2020supervised} &0.970 & 0.996 & 0.150 \\
      &            & \citet{uppaal2023fine} & 0.990 & 0.998 & 0.012 \\
      &            & Llama-7B LH  & 0.755 & 0.311 & 0.932 \\
      &            & Llama-7B LR & \textbf{1.000} & \textbf{1.000} & 0.001  \\
      &            & Mistral-7B LH  & 0.767 & 0.943 & 0.926 \\
      &            & Mistral-7B LR & 0.999 & 0.998 & 0.003 \\
      &            & Llama-13B LH & 0.773 & 0.332 & 0.919 \\
      &            & Llama-13B LR & \textbf{1.000} & \textbf{1.000} & \textbf{0.000} \\
\bottomrule
\specialrule{0em}{-3pt}{0pt}
\end{tabular}
\caption{Results of far OOD detection, utilizing the same experimental setup as described by \citet{uppaal2023fine}. Results for methods not originating from our work are cited directly from \citet{uppaal2023fine}. For the complete details, please see Appendix. \label{tab:farood}}
\end{table*}

\section{EXPERIMENTS}
In this section, we conduct a comprehensive evaluation across various scenarios, including far OOD, near OOD, spam detection, and QA, to demonstrate the effectiveness of our approach.

We adhere to the definitions of far OOD and near OOD as outlined by \citet{yang2022openood} in their work. Near OOD datasets exhibit only a semantic shift from the In-D datasets, whereas far OOD also encompasses a significant covariate (domain) shift. For far OOD evaluations, we designate two distinct datasets as In-D and OOD. For near OOD, we divide a single dataset into two groups: one serving as the In-D with certain classes and the other as OOD with a different set of classes.

Additionally, we demonstrate the capability of our proposed method in detecting OOD instances within the context of spam detection \citep{labonne2023spam} --- a practical application for our unsupervised OOD detection technique, especially where In-D labels are absent. We show that our method achieves commendable results in spam detection even without access to any spam data. Moreover, when spam data is available and we further finetune the OOD proxy distribution using this data, our results are competitive with state-of-the-art (SOTA) spam detection algorithms.

Finally, we evaluate our approach within a real question-answering (QA) context, utilizing MetaMath --- a Llama-2 model finetuned for math problem-solving, as described by \citet{yu2023metamath}. By implementing the likelihood-ratio-based criteria outlined in Section~\ref{sec:qa}, we find that for specific short questions, having the LLM provide an answer and subsequently applying a criterion that analyzes both the question and answer leads to consistently improved outcomes in identifying OOD questions.

\paragraph{Evaluation Metrics}
We employ AUROC (Area Under the Receiver Operating Characteristic curve), AUPR (Area Under the Precision-Recall curve), and FPR95 (False Positive Rate at 95\% True Positive Rate) as our evaluation metrics. These metrics are commonly utilized in evaluating the performance of OOD detection methods \citep{hendrycks2016baseline, yang2022openood}.

\begin{table*}[t]
\scriptsize
\setlength{\tabcolsep}{13pt}
\renewcommand{\arraystretch}{1.05}
\centering
\begin{tabular}{cclccc}
\specialrule{0em}{0pt}{-3pt}
\toprule
\multicolumn{1}{c}{\bf Dataset} & \multicolumn{1}{c}{\bf In-D Label}  &\multicolumn{1}{c}{\bf Model} &\multicolumn{1}{c}{\bf AUROC $\uparrow$} &\multicolumn{1}{c}{\bf AUPR $\uparrow$} &\multicolumn{1}{c}{\bf FPR95 $\downarrow$}\\
\midrule
\multirow{9}{*}{ROSTD}   & \multirow{8}{*}{No}    & \citet{gangal2020likelihood} & 0.981 & 0.958 & 0.077 \\
       &       & \citet{jin2022towards} & 0.990 & 0.973 & 0.041 \\
       &       & Llama-7B LH  & 0.960 & 0.890 & 0.168 \\
       &       & Llama-7B LR  & \textbf{0.994} & 0.984 & 0.023 \\
       &       & Mistral-7B LH  & 0.964 & 0.901 & 0.158 \\
       &       & Mistral-7B LR  & 0.992 & 0.978 & 0.033 \\
       &       & Llama-13B LH  & 0.965 & 0.905 & 0.166 \\
       &       & Llama-13B LR  & \textbf{0.994} & \textbf{0.988} & \textbf{0.018} \\
             \cmidrule(l){2-6}
       &  Yes  & \citet{podolskiy2021revisiting} & 0.998 & 0.994 & 0.008 \\
             \cmidrule(l){1-6}
\multirow{9}{*}{SNIPS}  & \multirow{8}{*}{No}    & \citet{gangal2020likelihood} & 0.955 & 0.903 & 0.192 \\
       &       & \citet{jin2022towards} & 0.963& 0.910 & 0.145 \\
       &       & Llama-7B LH  & 0.912 & 0.829 & 0.391 \\
       &       & Llama-7B LR  & 0.993 & 0.986 & 0.029 \\
       &       & Mistral-7B LH  & 0.912 & 0.819 & 0.417 \\
       &       & Mistral-7B LR  & 0.987 & 0.968 & 0.087 \\
       &       & Llama-13B LH  & 0.942 & 0.872 & 0.280 \\
       &       & Llama-13B LR  & \textbf{0.995} & \textbf{0.988} & \textbf{0.028} \\
             \cmidrule(l){2-6}
       &  Yes  & \citet{podolskiy2021revisiting} & 0.978 & 0.933 & 0.120 \\
             \cmidrule(l){1-6}
\multirow{9}{*}{CLINC150}  & \multirow{8}{*}{No}    & \citet{gangal2020likelihood} & 0.883 & 0.677 & 0.463 \\
       &       & \citet{jin2022towards} & 0.902 & 0.703 & 0.417 \\
       &       & Llama-7B LH  & 0.821 & 0.456 & 0.538 \\
       &       & Llama-7B LR  & \textbf{0.917} & \textbf{0.766} & \textbf{0.384} \\
       &       & Mistral-7B LH  & 0.823 & 0.454 & 0.540 \\
       &       & Mistral-7B LR  & 0.913 & 0.730 & 0.399 \\
       &       & Llama-13B LH  & 0.820 & 0.450 & 0.546 \\
       &       & Llama-13B LR  & 0.915 & 0.742 & 0.386 \\
             \cmidrule(l){2-6}
       &  Yes  & \citet{podolskiy2021revisiting} & 0.982 & 0.939 & 0.092 \\
\bottomrule
\specialrule{0em}{-3pt}{0pt}
\end{tabular}
\caption{
Results for near OOD detection. Since the experimental configurations in the studies by \citet{gangal2020likelihood}, \citet{podolskiy2021revisiting}, and \citet{jin2022towards} differ, we have replicated their methods and aligned the dataset splitting for consistency.\label{tab:nearood}}
\end{table*}

\subsection{Far OOD Detection}
For text data, detecting far OOD instances is relatively less challenging. \citet{uppaal2023fine} have illustrated that utilizing the latent distance from a pretrained RoBERTa model significantly addresses far OOD detection, especially when the 20 Newsgroups (20NG) dataset serves as the in-distribution. In this context, we present that our proposed method also attains nearly perfect performance under the same experimental conditions, as detailed in Table~\ref{tab:farood}. 

Note that, the notation `Model-XB LH/LR' is used, where `Model' can be either Llama \citep{touvron2023llama} or Mistral \citep{jiang2023mistral}, denoting two popular open-source large language models (LLMs). Here, `Llama' specifically refers to the Llama 2 version. The `XB' indicates the model's parameter count, either 7B or 13B. `LH' stands for likelihood, signifying the use of $p_\theta (x)$ as the criterion for OOD detection; `LR' denotes likelihood-ratio, referring to the employment of $S(x)$ as outlined in Equation (\ref{eq:lr}) for OOD identification. 

Examining Table~\ref{tab:farood}, we observe that for far OOD detection, the Llama-13B LR model nearly perfectly addresses the challenge across all the In-D OOD pairs for far OOD detection. 

\begin{table*}[t]
\scriptsize
\setlength{\tabcolsep}{13pt}
\renewcommand{\arraystretch}{1.05}
\centering
\begin{tabular}{cclccc}
\specialrule{0em}{0pt}{-3pt}
\toprule
\multicolumn{1}{c}{\bf Dataset} & \multicolumn{1}{c}{\bf Spam Data}  &\multicolumn{1}{c}{\bf Model} &\multicolumn{1}{c}{\bf AUROC $\uparrow$} &\multicolumn{1}{c}{\bf AUPR $\uparrow$} &\multicolumn{1}{c}{\bf FPR95 $\downarrow$}\\
\midrule
\multirow{14}{*}{SMS}   &  \multirow{4}{*}{No}  & Llama-7B LH & 0.960 & 0.699 & 0.088  \\
      &      & Llama-7B LR & 0.866 & 0.582 & 0.487  \\
      &      & Llama-13B LH & 0.957 & 0.689 & 0.093 \\
      &      & Llama-13B LR & 0.810 & 0.518 & 0.761 \\
\cmidrule(l){2-6}
      &  \multirow{10}{*}{Yes} & NB & 0.988 & 0.949 & 0.113 \\
      &      & Logistic & 0.985 & 0.946 & 0.124  \\
      &       & KNN &  0.863 & 0.830 & 0.811  \\
      &       & SVM &  0.997 & 0.980 & 0.024 \\
      &       & XGBoost & 0.918 & 0.873 & 0.676 \\
      &       & LightGBM & 0.978 & 0.921 & 0.103 \\
             \cmidrule(l){3-6}
      &       & RoBERTa & 0.997 & 0.988 & 0.004 \\
      &       & Spam-T5 & 0.985 & 0.959 & 0.082 \\
             \cmidrule(l){3-6}
      &       & Llama-7B LR & \textbf{1.000} & \textbf{1.000} & \textbf{0.000} \\
      &       & Llama-13B LR & 0.999 & 0.995 & \textbf{0.000} \\
\cmidrule(l){1-6}
\multirow{14}{*}{SpamAssassin}   &  \multirow{4}{*}{No}  & Llama-7B LH & 0.964 & 0.884 & 0.096 \\
      &      & Llama-7B LR & 0.960 & 0.935 & 0.296\\
      &      & Llama-13B LH & 0.956 & 0.897 & 0.169 \\
      &      & Llama-13B LR & 0.941 & 0.917 & 0.398 \\
\cmidrule(l){2-6}
      &  \multirow{10}{*}{Yes} & NB & 0.971 & 0.917 & 0.070 \\
      &      & Logistic & 0.992 & 0.986 & 0.029 \\
      &       & KNN & 0.931 & 0.935 & 0.578 \\
      &       & SVM & 0.990 & 0.983 & 0.046 \\
      &       & XGBoost & 0.994 & 0.989 & 0.019 \\
      &       & LightGBM & \textbf{1.000} & \textbf{0.999} & \textbf{0.000} \\
             \cmidrule(l){3-6}
      &       & RoBERTa & 0.999 & 0.997 & \textbf{0.000} \\
      &       & Spam-T5 & 0.996 & 0.994 & 0.012 \\
             \cmidrule(l){3-6}
      &       & Llama-7B LR & 0.998 & 0.996 & 0.005 \\
      &       & Llama-13B LR & 0.994 & 0.989 & 0.019 \\
\cmidrule(l){1-6}
\multirow{14}{*}{Enron}   &  \multirow{4}{*}{No}  & Llama-7B LH & 0.721 & 0.728 & 0.798 \\
      &      & Llama-7B LR & 0.991 & 0.989 & 0.043 \\
      &      & Llama-13B LH & 0.719 & 0.723 & 0.798 \\
      &      & Llama-13B LR & 0.992 & 0.990 & 0.035 \\
\cmidrule(l){2-6}
      &  \multirow{10}{*}{Yes} & NB & 0.992 & 0.991 & 0.035 \\
      &      & Logistic & 0.994 & 0.992 & 0.025 \\
      &       & KNN & 0.915 & 0.927 & 0.239 \\
      &       & SVM & 0.998 & 0.998 & 0.008 \\
      &       & XGBoost & 0.975 & 0.967 & 0.111 \\
      &       & LightGBM & 0.997 & 0.997 & 0.013 \\
             \cmidrule(l){3-6}
      &       & RoBERTa & \textbf{1.000} & \textbf{1.000} & 0.001 \\
      &       & Spam-T5 & \textbf{1.000} & \textbf{1.000} & 0.001 \\
             \cmidrule(l){3-6}
      &       & Llama-7B LR & 0.999 & 0.999 & 0.001 \\
      &       & Llama-13B LR & \textbf{1.000} & \textbf{1.000} & \textbf{0.000} \\
\bottomrule
\specialrule{0em}{-3pt}{0pt}
\end{tabular}
\caption{Results of spam detection. For the complete details, please see Appendix.\label{tab:spam}}
\end{table*}

\subsection{Near OOD Detection}
We select the ROSTD \citep{gangal2020likelihood}, SNIPS \citep{coucke2018snips}, and CLINC150 \citep{larson2019evaluation} datasets for our near OOD detection experiments. The ROSTD and CLINC150 datasets are specifically crafted for OOD detection and include designated classes representing OOD data from the same domain. The SNIPS dataset comprises user utterances distributed among seven intent classes, such as GetWeather and RateBook. As it does not inherently provide OOD utterances, we classify the GetWeather and BookRestaurant intents as OOD for the purpose of our experiments. It's noteworthy that this classification diverges from the one in the study by \citet{jin2022towards}, which does not explicitly detail their data splitting methodology.

Table~\ref{tab:nearood} demonstrates that the likelihood ratio between the pretrained Llama model and the finetuned Llama model yields the highest performance among unsupervised OOD detection methods. In the case of CLINC150, the supervised OOD detection method introduced by \citet{podolskiy2021revisiting} significantly surpasses our approach, a point that is further discussed in Section~\ref{sec:indlabels}.

\subsection{Spam Detection}
Given that the concept of unsupervised OOD detection aligns closely with spam detection, we evaluate our method using the spam detection benchmark introduced by \citet{labonne2023spam}. This benchmark includes four specific spam detection datasets: Ling-Spam Dataset \citep{sakkis2003memory}, SMS Spam Collection \citep{almeida2011contributions}, SpamAssassin Public Corpus, and Enron Email Dataset \citep{metsis2006spam}. It compares the performance of both traditional and deep learning-based binary classifiers. Our method, being rooted in OOD detection, requires only the non-spam (ham) data for finetuning the LLM. Table~\ref{tab:spam} indicates that our method, without any spam data, can still reliably identify spam. Furthermore, when spam data is available, we can finetune the OOD proxy using this data and apply the likelihood ratio between the two finetuned LLMs. This approach achieves performance that is competitive with the SOTA in spam detection.

\subsection{OOD Question Detection in QA Systems}

We test the effectiveness of the criterion we introduced in Section~\ref{sec:qa} within a QA scenario. Here, we employ MetaMath \citep{yu2023metamath}, law-chat and medicine-chat \citep{cheng2023adapting}, which are Llama2 models finetuned on specific domains. The objective in this QA context is to identify OOD questions that originate from domains outside the model's expertise. Filtering out OOD questions enhances the system's robustness. For this evaluation, we designate the test sets of the GSM8K \citep{cobbe2021training}, MATH \citep{hendrycks2021measuring}, casehold \citep{zheng2021does} and PubMedQA \citep{jin2019pubmedqa} datasets as In-D and use SQUAD \citep{rajpurkar2018know}, BoolQ \citep{clark2019boolq}, and PIQA \citep{bisk2020piqa} as OOD datasets. Table~\ref{tab:qa} indicates that the criterion $S_{a}$ consistently outperforms the other evaluated criteria, with all its AUROC values exceeding 0.5. This demonstrates its effectiveness in detecting OOD. Notably, $S_q$ exhibits subpar performance in most scenarios, underscoring the necessity of our proposed approach that generates an answer and formulates the criterion based on the question-and-answer pair.

\begin{table*}[t]
\scriptsize
\setlength{\tabcolsep}{11pt}
\renewcommand{\arraystretch}{1.05}
\centering
\begin{tabular}{ccclccc}
\specialrule{0em}{0pt}{-7pt}
\toprule
\multicolumn{1}{c}{\bf In-D} &\multicolumn{1}{c}{\bf OOD} &\multicolumn{1}{c}{\bf Model} &\multicolumn{1}{c}{\bf Criterion} & \multicolumn{1}{c}{\bf AUROC $\uparrow$} &\multicolumn{1}{c}{\bf AUPR (OOD) $\uparrow$} &\multicolumn{1}{c}{\bf FPR95 $\downarrow$}\\
\midrule
\multirow{12}{*}{MATH}  & \multirow{4}{*}{SQUAD 2.0} & \multirow{4}{*}{MetaMath-7B} & $S_q$ & 0.2139 & 0.2304 & 0.9474  \\
      &            &    & $S_a$ & 0.6384 & 0.3963 & 0.8916  \\
      &            &    & $S_{q,a}$& 0.6527 & 0.4477 & 0.8436  \\
      &            &    & $S_{a|q}$ & \textbf{0.7385} & \textbf{0.5305} & \textbf{0.7914}  \\
\cmidrule(l){2-7}
      & \multirow{4}{*}{BoolQ} & \multirow{4}{*}{MetaMath-7B} & $S_q$ & 0.1303 & 0.4472 & 0.9658  \\
      &            &    & $S_a$ & 0.6135 & 0.7111 & 0.8580  \\
      &            &    & $S_{q,a}$ & 0.6361 & 0.7474 & 0.8008  \\
      &            &    & $S_{a|q}$ & \textbf{0.7507} & \textbf{0.8266} & \textbf{0.6870}  \\
\cmidrule(l){2-7}
      & \multirow{4}{*}{PIQA}       & \multirow{4}{*}{MetaMath-7B} & $S_q$ & 0.9681 & 0.9902 & 0.0732  \\
      &            &    & $S_a$ & 0.9206 & 0.9775 & 0.1133  \\
      &            &    & $S_{q,a}$ & 0.9873 & \textbf{0.9962} & \textbf{0.0242}  \\
      &            &    & $S_{a|q}$ & \textbf{0.9876} & 0.9956 & 0.0812  \\
\cmidrule(l){1-7}
\multirow{12}{*}{casehold}  & \multirow{4}{*}{SQUAD 2.0} & \multirow{4}{*}{law-chat-7B} & $S_q$ & 0.2463 & 0.2039 & 0.9987  \\
      &            &    & $S_a$ & \textbf{0.9082} & \textbf{0.8425} & \textbf{0.3717}  \\
      &            &    & $S_{q,a}$& 0.2048 & 0.1954 & 0.9989  \\
      &            &    & $S_{a|q}$ & 0.3915 & 0.2080 & 0.9992  \\
\cmidrule(l){2-7}
      & \multirow{4}{*}{BoolQ} & \multirow{4}{*}{law-chat-7B} & $S_q$ & 0.3869 & 0.5249 & 0.9985  \\
      &            &    & $S_a$ & \textbf{0.8878} & \textbf{0.9222} & \textbf{0.4753}  \\
      &            &    & $S_{q,a}$ & 0.2692 & 0.4737 & 0.9988  \\
      &            &    & $S_{a|q}$ & 0.1307 & 0.3257 & 0.9996 \\
\cmidrule(l){2-7}
      & \multirow{4}{*}{PIQA}       & \multirow{4}{*}{law-chat-7B} & $S_q$ & 0.9241 & 0.9717 & 0.3064  \\
      &            &    & $S_a$ & \textbf{0.9917} & \textbf{0.9978} & \textbf{0.0078}  \\
      &            &    & $S_{q,a}$ & 0.8539 & 0.9440 & 0.4832  \\
      &            &    & $S_{a|q}$ & 0.0045 & 0.3743 & 0.9984  \\

\bottomrule
\specialrule{0em}{-3pt}{0pt}
\end{tabular}
\caption{OOD question detection in QA settings. For the complete details, please see Appendix. \label{tab:qa}}
\end{table*}

\section{DISCUSSIONS}

\paragraph{Nalisnick's Paradox in Language OOD Detection}
\label{sec:nalisnickllm}
Our far OOD detection experiments in Table~\ref{tab:farood} show that the AUROC for likelihood-based OOD detection methods can be exceptionally low. This finding echoes the phenomenon highlighted by \citet{nalisnick2018deep}, where language OOD data may unexpectedly exhibit higher likelihood values. Upon examining the characteristics of the datasets involved, it becomes apparent that the texts from the 20 Newsgroups (20NG) \citep{lang1995newsweeder} dataset are significantly longer than those from the comparative OOD datasets, such as SST-2 and RTE, especially in instances where the AUROC is notably low. This observation reveals a tendency among language models to assign higher likelihoods to shorter sentences, irrespective of their actual semantic content. It highlights the crucial importance of adopting the likelihood ratio rather than solely relying on raw likelihood values to enhance the accuracy of OOD detection.

\paragraph{The Effectiveness of In-D Labels}
\label{sec:indlabels}
In the near OOD detection experiments presented in Table~\ref{tab:nearood}, \citet{podolskiy2021revisiting}'s approach outperforms competing methods significantly. This superior performance may be attributed to the distinct class distribution across datasets. Specifically, the CLINC150 dataset comprises 150 In-D classes, in stark contrast to the SNIPS and ROSTD datasets, which offer a mere 7 (with only 5 for In-D) and 13 In-D classes, respectively. The substantially greater number of In-D classes in CLINC150 compared to the other datasets likely enhances \citet{podolskiy2021revisiting}'s method's ability to leverage the extensive class label information, thus yielding improved OOD detection results.

\paragraph{Some cases that LH is better than LR}
In the spam detection experiments detailed in Table~\ref{tab:spam}, particularly with data from the SMS and SpamAssassin datasets, we observe that without spam data, the likelihood (LH) method outperforms the likelihood ratio (LR). While LR generally shows superior and more consistent performance across most experiments --- as LH can exhibit extremely poor performance in certain cases, a point elaborated in Section~\ref{sec:nalisnickllm} --- there are specific instances where LH is more effective.

The rationale behind using large language models (LLMs) as an OOD proxy, as introduced in Section~\ref{sec:oodproxy}, is based on the assumption that OOD data deviates from domain-specific natural language content. However, spam messages in the SMS dataset often include intentionally misspelled words to circumvent detection mechanisms, thereby violating our natural language assumption. Similarly, the content from the SpamAssassin dataset, being highly structured in email and data transaction formats (header information), also diverges from typical natural language patterns.

Given these deviations from the natural language assumption, it is understandable why LH might outperform LR in these unique scenarios.

\section{RELATED WORKS}
\label{sec:relatedwork}
Building upon the OOD detection method using likelihood ratios introduced by \citet{ren2019likelihood}, \citet{gangal2020likelihood} suggested employing the likelihood ratio between two LSTM language models. In their approach, one model functions as a ``background model'' representing OOD data and is trained on random combinations from the vocabulary.

\citet{jin2022towards} used the likelihood ratio between a pretrained GPT-2 and a finetuned version of GPT-2 as
a baseline to compare with their proposed contrastive learning-based OOD detection method.
They showed that their proposed method can out perform the likelihood ratio based method.
However, they only used GPT-2 in their likelihood ratio baseline, which is very small both in terms of training data and the model size by today standard.

In comparison, our study provides a more comprehensive analysis of likelihood ratio between base models and fine-tuned models using much larger LLMs.
We show that leveraging these more advanced LLMs, likelihood ratio significantly outperforms results from \citet{jin2022towards} in OOD detection. 
Additionally, in the current era, accessing and sharing both pretrained and finetuned LLMs has become considerably easier through online platforms such as Hugging Face. Using likelihood ratio in LLMs for OOD detection is easy, accessible, and very effective.

\section{CONCLUDING REMARKS}
We revisit and validate the likelihood ratio between a pretrained LLM and its finetuned variant as a criterion for OOD detection across various scenarios, without the need for additional training. This LLM-based likelihood ratio, despite being very easy to implement, shows surprising empirical effectiveness in OOD detection, and more importantly, it enables us to build robust QA systems that are able to answer both general and domain-specific questions (i.e., using the likelihood ratio to determine which LLM should be used to answer the input question). We expect that our LLM-based likelihood ratio can benefit many other applications in the future.

\section*{ACKNOWLEDGEMENTS}
In the early stages of this work, AZ was supported by a personal grant from Mrs.~Yanshu Wu. In the middle and later stages, AZ was funded by the UKRI AI Hub in Generative Models (EP/Y028805/1). 
RB and TX were supported by the German Federal Ministry of Education and Research (BMBF): Tübingen AI Center, FKZ: 01IS18039A.
RB acknowledges funding by the German Research Foundation (DFG) for project 448588364 of the Emmy Noether Programme.
The authors thank the International Max Planck Research School for Intelligent Systems (IMPRS-IS) for supporting Tim Z. Xiao.

\bibliography{ref}


\onecolumn
\aistatstitle{Your Finetuned Large Language Model is Already a Powerful Out-of-distribution Detector: Supplementary Materials}

\begin{figure*}[htbp]
\begin{center}
\begin{tikzpicture}[scale=1, every node/.style={scale=1}]
\node[inner sep=0pt, anchor=north] (prompt) at (0,0)
    {\includegraphics[page=1, width=0.32\textwidth]{prompt.pdf}};
\node[inner sep=0pt, anchor=north] (prompt) at (5.5,0)
    {\includegraphics[page=2, width=0.32\textwidth]{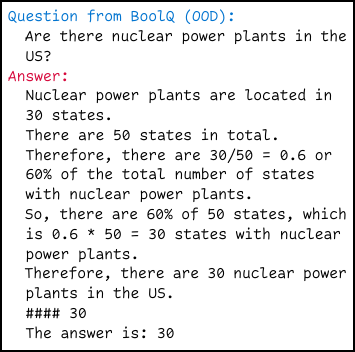}};
\node[inner sep=0pt, anchor=north] (prompt) at (11,0)
    {\includegraphics[page=3, width=0.32\textwidth]{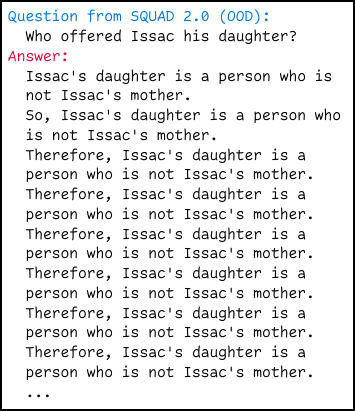}};
\end{tikzpicture}
\end{center}
\caption{Example question-answer sets produced by MetaMath-7B. The responses to In-D questions are accurate and logical. However, for OOD questions, MetaMath-7B generates unreasonable answers, responding to a straightforward query with unnecessary mathematical calculations or producing repetitive sentences with no useful information. \label{fig:qaexamplefull}}
\end{figure*}

\section{Finetuning Details}
For finetuning, with the exception of MetaMath \citep{yu2023metamath}, which is comprehensively finetuned from the Llama-2 model, we employ LoRA (Low-Rank Adaptation) as described by \citet{hu2021lora} for adjusting the models. Our aim is to demonstrate the general applicability of our method; therefore, all our LoRA finetuning follows a uniform parameter configuration. In accordance with prevalent practices, we finetune the parameters in the Q (query) and V (value) projections. We employ a LoRA rank of 16, a LoRA alpha of 32, and a LoRA dropout rate of 0.05. Given the varying sizes of datasets, we adjust the learning rate between $10^{-3}$ and $10^{-4}$, and the training typically spans 10 to 20 epochs. We select the model checkpoint with the best evaluation loss as our finetuned model. For detailed information, please refer to the code provided.

\section{The Use of AI Assistants in Research and Writing}
We assert that our use of AI assistants is strictly limited to revising the original text.

\section{Description of Computing Infrastructure}
All experiments were conducted using NVIDIA GPUs with varying memory capacities based on model size requirements. For models with parameter counts below 7B, we utilized NVIDIA V100 GPUs with 32GB VRAM and NVIDIA RTX 4090 GPUs with 24GB VRAM. For larger models with 13B parameters, we employed NVIDIA H100 GPUs with 80GB VRAM to accommodate the increased memory demands during training and inference.

\section{Full Experiment Results}

Full experiment results are detailed below. All the results are averaged over five runs.

\begin{table*}[ht]
\tiny
\begin{center}
\begin{tabular}{cclccc}
\toprule
\multicolumn{1}{c}{\bf In-D} &\multicolumn{1}{c}{\bf OOD} &\multicolumn{1}{c}{\bf Method} &\multicolumn{1}{c}{\bf AUROC $\uparrow$} &\multicolumn{1}{c}{\bf AUPR (OOD) $\uparrow$} &\multicolumn{1}{c}{\bf FPR95 $\downarrow$}\\
\midrule
\multirow{77}[10]{*}{20NG}  & \multirow{11}{*}{SST-2}     & \citet{zhou2021contrastive} & 0.978 & 0.865 & 0.015  \\
      &            & CE \citep{hendrycks2016baseline} & 0.981 & 0.942  & 0.087 \\
      &            & TAPT \citep{gururangan2020don} & 0.981 & 0.939 & 0.088 \\
      &            & SupCon \citep{khosla2020supervised} & 0.980 & 0.943 & 0.094 \\
      &            & \citet{uppaal2023fine} & \textbf{1.000} & 0.999  & \textbf{0.000}\\
      &            & Llama-7B LH & 0.008 & 0.541 & 0.999  \\
      &            & Llama-7B LR & \textbf{1.000} & \textbf{1.000} & \textbf{0.000} \\
      &            & Mistral-7B LH  & 0.008 & 0.541 & 1.000 \\
      &            & Mistral-7B LR & 0.995 & 0.999 & 0.009 \\
      &            & Llama-13B LH & 0.009 & 0.541 & 1.000  \\
      &            & Llama-13B LR & \textbf{1.000} & \textbf{1.000} & \textbf{0.000} \\
             \cmidrule(l){2-6}
      & \multirow{11}{*}{MNLI}       & \citet{zhou2021contrastive} & 0.964  & 0.978 & 0.224 \\
      &            & CE \citep{hendrycks2016baseline} & 0.968  & 0.989 & 0.166 \\
      &            & TAPT \citep{gururangan2020don} &0.964  & 0.988 & 0.175 \\
      &            & SupCon \citep{khosla2020supervised} &0.970  & 0.990 & 0.156 \\
      &            & \citet{uppaal2023fine} &\textbf{1.000}  & \textbf{1.000} & \textbf{0.000} \\
      
      &            & Llama-7B LH  & 0.020 & 0.119 & 0.999 \\
      &            & Llama-7B LR & \textbf{1.000} & \textbf{1.000}  & 0.001 \\
      &            & Mistral-7B LH  & 0.024 & 0.119
 & 0.999 \\
      &            & Mistral-7B LR & 0.996  & 0.996 & 0.008 \\
      &            & Llama-13B LH & 0.024 & 0.119 & 0.998  \\
      &            & Llama-13B LR & \textbf{1.000} & \textbf{1.000} & \textbf{0.000} \\
             \cmidrule(l){2-6}
      & \multirow{11}{*}{RTE}        & \citet{zhou2021contrastive} &0.956  & 0.860 & 0.312 \\
      &            & CE \citep{hendrycks2016baseline} &0.945  & 0.902 & 0.285 \\
      &            & TAPT \citep{gururangan2020don} &0.919 &  0.869 & 0.352 \\
      &            & SupCon \citep{khosla2020supervised} &0.952 &  0.914 & 0.248 \\
      &            & \citet{uppaal2023fine} & \textbf{1.000} &  0.999 & \textbf{0.000} \\
      &            & Llama-7B LH  & 0.063 & 0.443 & 0.998 \\
      &            & Llama-7B LR & \textbf{1.000} & \textbf{1.000} & 0.001 \\
      &            & Mistral-7B LH & 0.074 & 0.446 & 0.998 \\
      &            & Mistral-7B LR & 0.997 & 0.999 & 0.006  \\
      &            & Llama-13B LH & 0.070 & 0.445 & 0.997 \\
      &            & Llama-13B LR & \textbf{1.000} & \textbf{1.000} & \textbf{0.000}  \\
             \cmidrule(l){2-6}
      & \multirow{11}{*}{IMDB}       & \citet{zhou2021contrastive} & 0.969 & 0.996 & 0.144 \\
      &            & CE \citep{hendrycks2016baseline} &0.961 & 0.995 & 0.206 \\
      &            & TAPT \citep{gururangan2020don} & 0.965 & 0.995 & 0.159 \\
      &            & SupCon \citep{khosla2020supervised} &0.970 & 0.996 & 0.150 \\
      &            & \citet{uppaal2023fine} & 0.990 & 0.998 & 0.012 \\
      &            & Llama-7B LH  & 0.755 & 0.311 & 0.932 \\
      &            & Llama-7B LR & \textbf{1.000} & \textbf{1.000} & 0.001  \\
      &            & Mistral-7B LH  & 0.767 & 0.943 & 0.926 \\
      &            & Mistral-7B LR & 0.999 & 0.998 & 0.003 \\
      &            & Llama-13B LH & 0.773 & 0.332 & 0.919 \\
      &            & Llama-13B LR & \textbf{1.000} & \textbf{1.000} & \textbf{0.000} \\
             \cmidrule(l){2-6}
      & \multirow{11}{*}{Multi30K}       & \citet{zhou2021contrastive} & 0.980 & 0.888 & 0.005 \\
      &            & CE \citep{hendrycks2016baseline} &0.962 & 0.920 & 0.175 \\
      &            & TAPT \citep{gururangan2020don} &0.956 & 0.922 & 0.167 \\
      &            & SupCon \citep{khosla2020supervised} &0.955 & 0.918 & 0.201 \\
      &            & \citet{uppaal2023fine} & \textbf{1.000} & \textbf{1.000} & \textbf{0.000} \\
      &            & Llama-7B LH  & 0.002 & 0.470 & 1.000 \\
      &            & Llama-7B LR & \textbf{1.000} & \textbf{1.000} & \textbf{0.000} \\
      &            & Mistral-7B LH  & 0.002 & 0.470 & 1.000  \\
      &            & Mistral-7B LR & 0.995 & 0.998 & 0.008 \\
      &            & Llama-13B LH & 0.002 & 0.470 & 1.000  \\
      &            & Llama-13B LR & \textbf{1.000} & \textbf{1.000} & \textbf{0.000} \\
             \cmidrule(l){2-6}
      & \multirow{11}{*}{NewsCategory}        & \citet{zhou2021contrastive} &0.955 & 0.969 & 0.383 \\
      &            & CE \citep{hendrycks2016baseline} &0.957 & 0.984 & 0.234 \\
      &            & TAPT \citep{gururangan2020don} &0.947 & 0.981 & 0.243 \\
      &            & SupCon \citep{khosla2020supervised} &0.962 & 0.986 & 0.219 \\
      &            & \citet{uppaal2023fine} & \textbf{1.000} & \textbf{1.000} & \textbf{0.000} \\
      &            & Llama-7B LH  & 0.014 & 0.128 & 1.000 \\
      &            & Llama-7B LR & \textbf{1.000} & \textbf{1.000} & 0.001 \\
      &            & Mistral-7B LH  & 0.019 & 0.129 & 1.000 \\
      &            & Mistral-7B LR & 0.997 & 0.997 & 0.006 \\
      &            & Llama-13B LH & 0.017 & 0.128 &  0.999  \\
      &            & Llama-13B LR & \textbf{1.000} & \textbf{1.000} & \textbf{0.000} \\
      \cmidrule(l){2-6}
      & \multirow{11}{*}{CLINC150}       & \citet{zhou2021contrastive} &0.988 & 0.870 & 0.005 \\
      &            & CE \citep{hendrycks2016baseline} &0.964 & 0.844 & 0.189 \\
      &            & TAPT \citep{gururangan2020don} &0.959 & 0.830 & 0.213 \\
      &            & SupCon \citep{khosla2020supervised} &0.957 & 0.821 & 0.230 \\
      &            & \citet{uppaal2023fine} & \textbf{1.000} & \textbf{1.000} & \textbf{0.000} \\
      &            & Llama-7B LH  & 0.001 & 0.661 & 1.000 \\
      &            & Llama-7B LR & \textbf{1.000} & \textbf{1.000} & \textbf{0.000} \\
      &            & Mistral-7B LH & 0.001 & 0.661 & 1.000 \\
      &            & Mistral-7B LR & 0.995 & \textbf{1.000} & 0.008 \\
      &            & Llama-13B LH & 0.001 & 0.661 & 1.000  \\
      &            & Llama-13B LR & \textbf{1.000} & \textbf{1.000} & \textbf{0.000} \\
\bottomrule
\end{tabular}
\end{center}
\caption{Results of far OOD detection, utilizing the same experimental setup as described by \citet{uppaal2023fine}. Results for methods not originating from our work are cited directly from \citet{uppaal2023fine}. \label{apptab:farood}}
\end{table*}

\begin{table*}[t]
\tiny
\begin{center}
\begin{tabular}{cclccc}
\toprule
\multicolumn{1}{c}{\bf Dataset} & \multicolumn{1}{c}{\bf Spam Data}  &\multicolumn{1}{c}{\bf Model} &\multicolumn{1}{c}{\bf AUROC $\uparrow$} &\multicolumn{1}{c}{\bf AUPR $\uparrow$} &\multicolumn{1}{c}{\bf FPR95 $\downarrow$}\\
\midrule
Ling  &  No  & Llama-7B LH & 0.552 & 0.215 & 0.934 \\
      &      & Llama-7B LR & 0.967 & 0.858 & 0.174\\
      &      & Llama-13B LH & 0.534 & 0.182 & 0.929\\
      &      & Llama-13B LR & 0.933 & 0.746 & 0.336 \\
\cmidrule(l){2-6}
& Yes & NB & 1.000 & 1.000 & 0.000 \\
& & Logistic & 1.000 & 1.000 & 0.000 \\
& & KNN & 0.968 & 0.932 & 0.021 \\
& & SVM & 1.000 & 1.000 & 0.000 \\
& & XGBoost & 0.995 & 0.973 & 0.017 \\
& & LightGBM & 0.997 & 0.979 & 0.008 \\
\cmidrule(l){3-6}
& & RoBERTa & 1.000 & 1.000 & 0.000 \\
& & Spam-T5 & 1.000 & 1.000 & 0.000 \\
             \cmidrule(l){3-6}
      &       & Llama-7B LR & 0.998 & 0.993 & 0.008 \\
      &       & Llama-13B LR & 0.997 & 0.988 & 0.012 \\
\cmidrule(l){1-6}
SMS   &  No  & Llama-7B LH & 0.960 & 0.699 & 0.088  \\
      &      & Llama-7B LR & 0.866 & 0.582 & 0.487  \\
      &      & Llama-13B LH & 0.957 & 0.689 & 0.093 \\
      &      & Llama-13B LR & 0.810 & 0.518 & 0.761 \\
\cmidrule(l){2-6}
      &  Yes & NB & 0.988 & 0.949 & 0.113 \\
      &      & Logistic & 0.985 & 0.946 & 0.124  \\
      &       & KNN &  0.863 & 0.830 & 0.811  \\
      &       & SVM &  0.997 & 0.980 & 0.024 \\
      &       & XGBoost & 0.918 & 0.873 & 0.676 \\
      &       & LightGBM & 0.978 & 0.921 & 0.103 \\
             \cmidrule(l){3-6}
      &       & RoBERTa & 0.997 & 0.988 & 0.004 \\
      &       & Spam-T5 & 0.985 & 0.959 & 0.082 \\
             \cmidrule(l){3-6}
      &       & Llama-7B LR & 1.000 & 1.000 & 0.000 \\
      &       & Llama-13B LR & 0.999 & 0.995 & 0.000 \\
\cmidrule(l){1-6}
SpamAssassin  &  No  & Llama-7B LH & 0.964 & 0.884 & 0.096 \\
      &      & Llama-7B LR & 0.960 & 0.935 & 0.296\\
      &      & Llama-13B LH & 0.956 & 0.897 & 0.169 \\
      &      & Llama-13B LR & 0.941 & 0.917 & 0.398 \\
\cmidrule(l){2-6}
      &  Yes & NB & 0.971 & 0.917 & 0.070 \\
      &      & Logistic & 0.992 & 0.986 & 0.029 \\
      &       & KNN & 0.931 & 0.935 & 0.578 \\
      &       & SVM & 0.990 & 0.983 & 0.046 \\
      &       & XGBoost & 0.994 & 0.989 & 0.019 \\
      &       & LightGBM & 1.000 & 0.999 & 0.000 \\
             \cmidrule(l){3-6}
      &       & RoBERTa & 0.999 & 0.997 & 0.000 \\
      &       & Spam-T5 & 0.996 & 0.994 & 0.012 \\
             \cmidrule(l){3-6}
      &       & Llama-7B LR & 0.998 & 0.996 & 0.005 \\
      &       & Llama-13B LR & 0.994 & 0.989 & 0.019 \\
\cmidrule(l){1-6}
Enron  &  No  & Llama-7B LH & 0.721 & 0.728 & 0.798 \\
      &      & Llama-7B LR & 0.991 & 0.989 & 0.043 \\
      &      & Llama-13B LH & 0.719 & 0.723 & 0.798 \\
      &      & Llama-13B LR & 0.992 & 0.990 & 0.035 \\
\cmidrule(l){2-6}
      &  Yes & NB & 0.992 & 0.991 & 0.035 \\
      &      & Logistic & 0.994 & 0.992 & 0.025 \\
      &       & KNN & 0.915 & 0.927 & 0.239 \\
      &       & SVM & 0.998 & 0.998 & 0.008 \\
      &       & XGBoost & 0.975 & 0.967 & 0.111 \\
      &       & LightGBM & 0.997 & 0.997 & 0.013 \\
             \cmidrule(l){3-6}
      &       & RoBERTa & 1.000 & 1.000 & 0.001 \\
      &       & Spam-T5 & 1.000 & 1.000 & 0.001 \\
             \cmidrule(l){3-6}
      &       & Llama-7B LR & 0.999 & 0.999 & 0.001 \\
      &       & Llama-13B LR & 1.000 & 1.000 & 0.000 \\
\bottomrule
\end{tabular}
\end{center}
\caption{Results of spam detection. \label{apptab:spam}}
\end{table*}

\begin{table*}[t]
\tiny
\begin{center}
\begin{tabular}{ccllccc}
\toprule
\multicolumn{1}{c}{\bf In-D} &\multicolumn{1}{c}{\bf OOD} &\multicolumn{1}{c}{\bf Model} &\multicolumn{1}{c}{\bf Criterion} & \multicolumn{1}{c}{\bf AUROC $\uparrow$} &\multicolumn{1}{c}{\bf AUPR (OOD) $\uparrow$} &\multicolumn{1}{c}{\bf FPR95 $\downarrow$}\\
\midrule
GSM8K  & SQUAD 2.0 & MetaMath-7B & $S_q$ & 0.1116 & 0.0546 & 0.9894  \\
      &            &    & $S_a$ & 0.5463 & 0.0979 & 0.9947  \\
      &            &    & $S_{q,a}$ & 0.5363 & 0.0959 & 0.9924  \\
      &            &    & $S_{a|q}$ & \textbf{0.6877} & \textbf{0.1376} & \textbf{0.9704}  \\
\cmidrule(l){3-7}
       &           & MetaMath-13B & $S_q$ & 0.0519 & 0.0524 & 0.9955  \\
      &            &    & $S_a$ & 0.4958 & 0.0891 & 0.9955  \\
      &            &    & $S_{q,a}$ & 0.4017 & 0.0764 & 0.9947  \\
      &            &    & $S_{a|q}$ & \textbf{0.6144} & \textbf{0.1136} & \textbf{0.9765}  \\
\cmidrule(l){2-7}
      & BoolQ & MetaMath-7B & $S_q$ & 0.0538 & 0.1618 & 0.9955  \\
      &            &    & $S_a$ & 0.5045 & 0.2616 & 0.9932  \\
      &            &    & $S_{q,a}$ & 0.4797 & 0.2536 & 0.9879  \\
      &            &    & $S_{a|q}$ & \textbf{0.7156} & \textbf{0.4041} & \textbf{0.9310}  \\
             \cmidrule(l){3-7}
       &           & MetaMath-13B & $S_q$ & 0.1008 & 0.1659 & 0.9924  \\
      &            &    & $S_a$ & 0.5507 & 0.2861 & 0.9833  \\
      &            &    & $S_{q,a}$ & 0.4778 & 0.2530 & 0.9795  \\
      &            &    & $S_{a|q}$ & \textbf{0.6967} & \textbf{0.3886} & \textbf{0.9303}  \\
\cmidrule(l){2-7}
      & PIQA       & MetaMath-7B & $S_q$ & 0.9762 & 0.9779 & 0.0735  \\
      &            &    & $S_a$ & 0.9612 & 0.9746 & 0.0569  \\
      &            &    & $S_{q,a}$ & \textbf{0.9975} & \textbf{0.9983} & \textbf{0.0038}  \\
      &            &    & $S_{a|q}$ & 0.9944 & 0.9944 & 0.0099  \\
             \cmidrule(l){3-7}
       &           & MetaMath-13B & $S_q$ & 0.9900 & 0.9906 & 0.0318  \\
      &            &    & $S_a$ & 0.8879 & 0.9331 & 0.1122  \\
      &            &    & $S_{q,a}$ & \textbf{1.0000} & 0.9999 & \textbf{0.0000}  \\
      &            &    & $S_{a|q}$ & \textbf{1.0000} & \textbf{1.0000} & \textbf{0.0000}  \\
\cmidrule(l){1-7}
MATH  & SQUAD 2.0 & MetaMath-7B & $S_q$ & 0.2139 & 0.2304 & 0.9474  \\
      &            &    & $S_a$ & 0.6384 & 0.3963 & 0.8916  \\
      &            &    & $S_{q,a}$& 0.6527 & 0.4477 & 0.8436  \\
      &            &    & $S_{a|q}$ & \textbf{0.7385} & \textbf{0.5305} & \textbf{0.7914}  \\
\cmidrule(l){3-7}
       &           & MetaMath-13B & $S_q$ & 0.1880 & 0.2232 & 0.9460  \\
      &            &    & $S_a$ & 0.6098 & 0.3841 & 0.8890  \\
      &            &    & $S_{q,a}$ & 0.5628 & 0.3649 & 0.8882  \\
      &            &    & $S_{a|q}$ & \textbf{0.6786} & \textbf{0.4737} & \textbf{0.8166}  \\
\cmidrule(l){2-7}
      & BoolQ & MetaMath-7B & $S_q$ & 0.1303 & 0.4472 & 0.9658  \\
      &            &    & $S_a$ & 0.6135 & 0.7111 & 0.8580  \\
      &            &    & $S_{q,a}$ & 0.6361 & 0.7474 & 0.8008  \\
      &            &    & $S_{a|q}$ & \textbf{0.7507} & \textbf{0.8266} & \textbf{0.6870}  \\
             \cmidrule(l){3-7}
       &           & MetaMath-13B & $S_q$ & 0.2612 & 0.5223 & 0.9304  \\
      &            &    & $S_a$ & 0.6488 & 0.7474 & 0.8148  \\
      &            &    & $S_{q,a}$ & 0.6384 & 0.7521 & 0.7818  \\
      &            &    & $S_{a|q}$ & \textbf{0.7350} & \textbf{0.8191} & \textbf{0.6854}  \\
\cmidrule(l){2-7}
      & PIQA       & MetaMath-7B & $S_q$ & 0.9681 & 0.9902 & 0.0732  \\
      &            &    & $S_a$ & 0.9206 & 0.9775 & 0.1133  \\
      &            &    & $S_{q,a}$ & 0.9873 & \textbf{0.9962} & \textbf{0.0242}  \\
      &            &    & $S_{a|q}$ & \textbf{0.9876} & 0.9956 & 0.0812  \\
             \cmidrule(l){3-7}
       &           & MetaMath-13B & $S_q$ & 0.9795 & 0.9938 & 0.0452  \\
      &            &    & $S_a$ & 0.8495 & 0.9572 & 0.1627  \\
      &            &    & $S_{q,a}$ & \textbf{0.9897} & \textbf{0.9969} & \textbf{0.0198}  \\
      &            &    & $S_{a|q}$ & 0.9626 & 0.9895 & 0.0484  \\
\bottomrule
\end{tabular}
\end{center}
\caption{Outcomes of OOD question detection in QA settings. This table continues in Table~\ref{apptab:qa2}. \label{apptab:qa}}
\end{table*}

\begin{table*}[t]
\tiny
\begin{center}
\begin{tabular}{ccllccc}
\toprule
\multicolumn{1}{c}{\bf In-D} &\multicolumn{1}{c}{\bf OOD} &\multicolumn{1}{c}{\bf Model} &\multicolumn{1}{c}{\bf Criterion} & \multicolumn{1}{c}{\bf AUROC $\uparrow$} &\multicolumn{1}{c}{\bf AUPR (OOD) $\uparrow$} &\multicolumn{1}{c}{\bf FPR95 $\downarrow$}\\
\midrule
\multirow{12}{*}{casehold}  & \multirow{4}{*}{SQUAD 2.0} & \multirow{4}{*}{law-chat-7B} & $S_q$ & 0.2463 & 0.2039 & 0.9987  \\
      &            &    & $S_a$ & \textbf{0.9082} & \textbf{0.8425} & \textbf{0.3717}  \\
      &            &    & $S_{q,a}$& 0.2048 & 0.1954 & 0.9989  \\
      &            &    & $S_{a|q}$ & 0.3915 & 0.2080 & 0.9992  \\
\cmidrule(l){2-7}
      & \multirow{4}{*}{BoolQ} & \multirow{4}{*}{law-chat-7B} & $S_q$ & 0.3869 & 0.5249 & 0.9985  \\
      &            &    & $S_a$ & \textbf{0.8878} & \textbf{0.9222} & \textbf{0.4753}  \\
      &            &    & $S_{q,a}$ & 0.2692 & 0.4737 & 0.9988  \\
      &            &    & $S_{a|q}$ & 0.1307 & 0.3257 & 0.9996 \\
\cmidrule(l){2-7}
      & \multirow{4}{*}{PIQA}       & \multirow{4}{*}{law-chat-7B} & $S_q$ & 0.9241 & 0.9717 & 0.3064  \\
      &            &    & $S_a$ & \textbf{0.9917} & \textbf{0.9978} & \textbf{0.0078}  \\
      &            &    & $S_{q,a}$ & 0.8539 & 0.9440 & 0.4832  \\
      &            &    & $S_{a|q}$ & 0.0045 & 0.3743 & 0.9984  \\
\cmidrule(l){1-7}
\multirow{12}{*}{PubMedQA}  & \multirow{4}{*}{SQUAD 2.0} & \multirow{4}{*}{medicine-chat-7B} & $S_q$ & \textbf{0.9024} & 0.2729 & 0.4300  \\
      &            &    & $S_a$ & 0.7939 & \textbf{0.6476} & \textbf{0.3660}  \\
      &            &    & $S_{q,a}$& 0.8967 & 0.3959 & 0.4360  \\
      &            &    & $S_{a|q}$ & 0.5756 & 0.1869 & 0.8400 \\
\cmidrule(l){2-7}
      & \multirow{4}{*}{BoolQ} & \multirow{4}{*}{medicine-chat-7B} & $S_q$ & \textbf{0.9266} & 0.5906 & 0.3980  \\
      &            &    & $S_a$ & 0.8009 & \textbf{0.7057} & \textbf{0.3800} \\
      &            &    & $S_{q,a}$ & 0.9066 & 0.6281 & 0.4440 \\
      &            &    & $S_{a|q}$ & 0.3515 & 0.2356 & 0.8440 \\
\cmidrule(l){2-7}
      & \multirow{4}{*}{PIQA}       & \multirow{4}{*}{medicine-chat-7B} & $S_q$ & \textbf{0.9998} & \textbf{0.9994} & \textbf{0.0000}  \\
      &            &    & $S_a$ & 0.8606 & 0.8439 & 0.3000  \\
      &            &    & $S_{q,a}$ & 0.9995 & 0.9984 & 0.0020  \\
      &            &    & $S_{a|q}$ & 0.2471 & 0.2892 & 0.8420 \\
\bottomrule
\end{tabular}
\end{center}
\caption{Outcomes of OOD question detection in QA settings (continued).\label{apptab:qa2}}
\end{table*}

\end{document}